# Learning to Navigate in Indoor Environments:
# from Memorizing to Reasoning

Liulong Ma, Yanjie Liu*, Jiao Chen and Dong Jin

*Abstract*— Autonomous navigation is an essential capability of smart mobility for mobile robots. Traditional methods require the environment map to plan a collision-free path in workspace. Deep reinforcement learning (DRL) is a promising technique to realize the autonomous navigation task without a map, with which deep neural network can fit the mapping from observation to reasonable action through explorations. It should not only memorize the trained target, but more importantly, the planner can reason out the path to unseen goal. We proposed a new motion planner based on deep reinforcement learning that can arrive at new targets that have not been trained before in the indoor environment with RGB image and odometry only. The model has a structure of stacked Long Short-Term memory (LSTM). Finally, experiments were implemented in both simulated and real environments. The source code is available: https://github.com/marooncn/navbot.

## I. INTRODUCTION

Autonomous navigation is the core ability of mobile robot. It can be roughly described as the ability of the robot to arrive at the target without collision with any obstacles. Traditionally, navigation task is based on the metric map and precise localization to plan a collision-free path in workplace. D-star and Probabilistic Roadmap (PRM) method are two classic algorithms of map-based robot navigation. D-star algorithm generalizes the occupancy grid to a cost map and finds the path which minimizes the total cost of travel [1]. PRM method is a probabilistic algorithm, it sparsely samples the world map, in which process it finds *N* random points lie in free space, and then these points including the start and goal point are connected to these neighbours by straight line paths that do not cross any obstacles so as to create a network, and the traversable path can be found in this network [2]. These methods are useful for real-world applications but they must rely on the environment map, which is often represented as the occupancy grid, and paths are planned without a cognitive process.

In contrast, biological systems can navigate efficiently in previously unseen environments, building up internal representations of these environments in the process. Such internal representations are of central importance to Artificial Intelligence [3]. It is an important issue for mobile robots and intelligent vehicles to learn to build up such internal representations like animals. Mobile robots do not know how to arrive at the target position with given information in the initial and they need to learn from the experiences. As experiences increases, they can build up the internal representations and navigate to a new target or even navigate efficiently in a new environment.

Reinforcement Learning (RL) is such a technology that the agent learns from interactions with the environment and the corresponding feedback. Mathematically, RL chooses a serious of actions according to states in order to maximize the total rewards in an episode. It either chooses polices directly or evaluates the value of each action then uses ε-greedy or other methods to chooses actions. The former is called policy-based RL and the latter is value-based RL. Reinforcement learning has a main advantage that is independent of human-labeling. It is a useful way to learn control polices without referencing the ground-truth. Deep reinforcement learning (DRL) enhances RL by using deep neural network as more powerful non-linear function approximators. It has been successfully in video games [4], Go [5], robotic manipulation [6] and other fields [7]. It is a promising technology to help the vehicles to learn to navigate without a map [8].

## II. RELATED WORK

### A. DRL algorithms

Deep Q-Network (DQN) [4] is a first well-known DRL algorithm that enhances Q-learning algorithms with deep neural network and experience replay. DNN fits the mapping function much better and experience replay breaks the relevance of the data. The network outputs evaluated state-action value (called Q-value) of each action. Then the action is chosen by ε-greedy method to interact with environment. The network is updated by the mean square error between target Q-value and evaluated Q-value, making the evaluation more and more accurate. There are many improved algorithms of DQN about experience replay [9], target Q-value [10], network structure [11] and other aspects. Rainbow integrates all these improvements and gets much better performance than DQN in Atari games [12]. DQN is value-based and can only deal with discrete and low-dimensional actions. Deep deterministic policy gradient (DDPG) [13] and normalized normal function (NAF) [14] are two continuous variants of DQN. DDPG deploys the actor-critic framework, in which the actor outputs actions in

* This research has been sponsored by the National Key R&D Program of China(No. 2017YFB1303801), the Provincial Funding for National Key R&D Program(Task)(No. GX18A011), the Self-Planned Task (No. SKLRS201813B) of State Key Laboratory of Robotics and System (HIT).

Liulong Ma, Jiao Chen, and Yanjie Liu are with the State Key Laboratory of Robotics and System, Harbin Institute of Technology, China. E-mail: LiulongMa@outlook.com, yjliu@hit.edu.cn, Jeffery-Chen@outlook.com, hit_jindong@163.com.

continuous space and the critic evaluates its value. NAF represents the Q-value function in such a way that the best policy can be easily determined.

Policy gradient algorithms are much suitable for high-dimensional continuous control problems and they directly maximize the expected sum of rewards. But these algorithms have high variance and they are sensitive to the setting of hyperparameters. Several approaches have been proposed to make policy gradient algorithms more robust. Trust Region Policy Optimization (TRPO) is such an algorithm that it maximizes an objective function called surrogate objective subject to a constraint on the size of the policy update [15]. Proximal Policy Optimization (PPO) is a first-order algorithm that emulates the monotonic improvement of TRPO [16]. PPO strikes a balance between performance and generalization. It is currently the best comprehensive performance algorithm and is the default DRL algorithm of OpenAI.

### B. Mapless Navigation

Traditional mapless navigation is often called reactive navigation. Reactive navigation is just based on simple rules, such as Braitenberg vehicle and simple automata. Braitenberg vehicle directly connects motors with sensors and it moves in the plane to seek the maxima of a scalar field such as light intensity [17]. Simple automata perform goal seeking in the presence of obstacles. The strategy is always quite simple, for example, in a method the robot moves along a straight line towards its goal and if it encounters an obstacle it moves around the obstacle until it goes back to the original line [18]. These methods have no explicit internal representations of the environment and the walking path is clearly not optimal.

DRL-based navigation tries to solve the problem of mapless navigation with DRL algorithms. The agent learns to build up the internal representations from explorations. Broadly speaking, navigation tasks are divided into three levels. In the first level the robot just needs to go from the start point to a specified target and has robustness to the random noise. The second and third levels require the robot to have the ability of reasoning. In the second level, the robot needs to arrive at unseen targets in the same environment in which the agent is trained and in the third level robots need to navigate to any targets autonomously in any environments. The third level is close to the ability of animals and it is the ultimate goal of mapless navigation.

Recent works mainly pay attention to improving the sample-efficient of the first level tasks and to addressing the second level tasks. Dosovitskiy et al. [19] improved the sample-efficient by predicting the future. Zhu et al. [20] input the observation and the target image together into a deep siamese actor-critic model to navigate the robot in indoor scenes. Khan et al. [21] enhanced actor-critic with auxiliary rewards and memory augmented network. Ma et al. [22] decoupled the feature extraction module from DRL network. For the second level tasks, the network always equips with short-term memories to memorize the environment [23, 24, 25]. More advanced memory mechanisms that support the construction of rich internal representations of the agent's environment are also proposed [26, 27, 28, 29].

## III. IMPLEMENTATION OF MOTION PLANNER

### A. Problem Definition

The task we addressed in this article is to train the agent in indoor environments to realize autonomous navigation. The agent can not only memorize the trained target, but more importantly, it can reason out the path to unseen goal. The mobile robot is only equipped with visual sensor and encoders, taking RGB image and its position relative to the target as input. The previous data is also useful, for example the velocity in the last time step contains the motion information. We try to find such a mapping:

$$v_t = f(x_t, g_t, a_{t-1}) \quad (1)$$

where $x_t$ is the RGB image from visual sensor, $g_t$ is the relative position and $a_{t-1}$ is the necessary information from the last time step. The mapping function is fitted by neural network. A simple neural network structure is end-to-end that takes all input together. But it doesn't work at all because of the gaint difference in dimensions. It's important to design a reasonable network structure.

In the first level task of navigation, mobile robot just remembers the path from the start to specified target, and has robustness against random noise. In our previous work [22], RGB image is encoded into 32-dim features and these features are input to fully-connected layers together with target and motion information. The nodes in fully-connected layers can not only memorize the specified path, but also reason out the path to new target if the visual information is sparse distance to obstacles [30]. In our case, the monocular image contains much more information, and the depth needs to be inferred from adjacent consecutive frames.

### B. Stacked LSTM

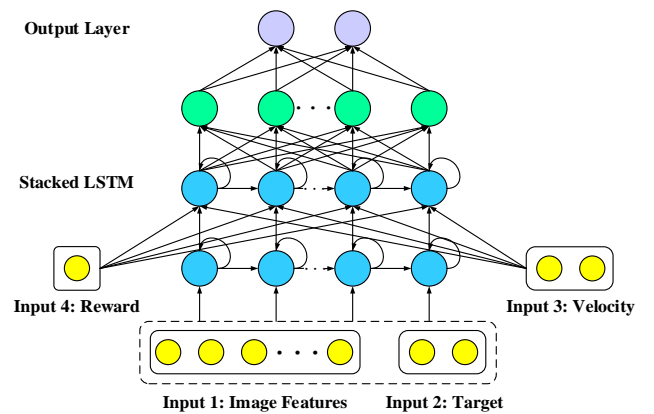

Figure 1. Stacked LSTM Structure

Convolutional neural network (CNN) is often applied to analyzing visual imagery and it has been successfully applied in image classification and medical image analysis. It is deployed to extract features of raw images. The extracted features are then input to Long Short-Term Memory (LSTM). LSTM is an efficient recurrent neural network architecture, dealing with the serious data. It can help to memorize the navigation environment and infer the depth from adjacent consecutive frames.

In our experiments, the single LSTM layer is always not better than stacked LSTM structure, in which the output of the LSTM layer is followed by the other LSTM layer. We proposed a motion planner using stacked LSTM. In the first LSTM layer, image features and the relative position of the target are taken as input. The output of this layer is then input to the following LSTM layer together with the reward and velocity in the last time step. The reward and velocity in the last time step help agent to understand this task [25, 31], so $a_{t-1} = \{v_{t-1}, r_{t-1}\}$. The structure of stacked LSTM is shown in Figure 1. Each LSTM layer in stacked LSTM has 256 units. The output of stacked LSTM then is connected to a fully-connected network to learn policy.

### C. Network Structure

The whole network structure of the proposed motion planner is shown in Figure 2.

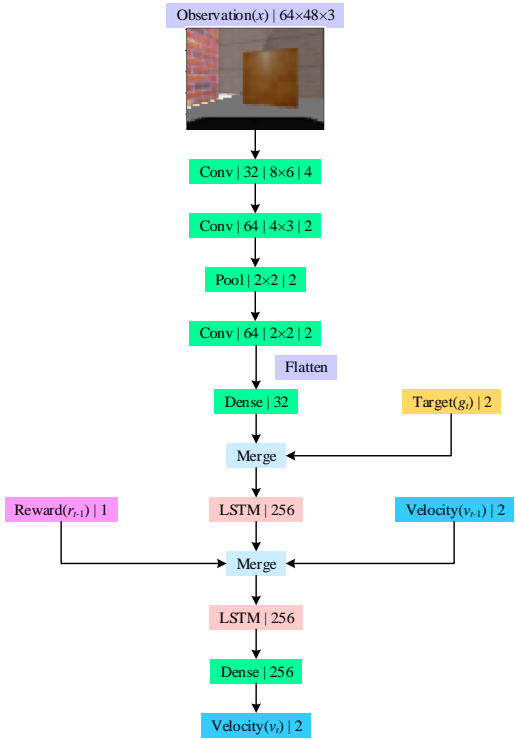

Figure 2. Network structure. Every convolutional layer is represented by its type, channel size, kernel size and stride size. Other layers are represented by their types and dimensions.

The observation $x_t$ is the raw image with $64 \times 48 \times 3$ dimension and $g_t$ is the relative position in polar coordinates (distance and angle) with respect to the mobile robot coordinate frame. In this network, a three-layer convolutional framework is used to perceive the indoor environment for mobile robots. Then the output of convolutional framework is connected to a fully-connected layer with 32 nodes to produce 32-dim features. Then the image features are input to stacked LSTM together with $g_t$ as illustrated in the previous part. After the last fully-connected neural network with 256 nodes, the input is transferred to the linear and angular velocity of the mobile robot. We choose 0.3 m/s as the max linear velocity and 1 rad/s as the max angular velocity. So the linear velocity is constrained in (0, 0.3) and the angular velocity is constrained in (-1, 1). The sigmoid function and tanh function are two corresponding activation functions of output layer.

### D. Motion Planner

The proposed motion planner uses PPO to update the parameters in the network:

$$\theta \leftarrow \arg\max_{\theta} \mathop{E}_{s,a \sim \pi_{\theta_k}} [L(s,a,\theta_k,\theta)] \quad (2)$$

L is given by:

$$L(s,a,\theta_k,\theta) = \min(r(\theta)A(s,a), clip(r(\theta),1-\varepsilon,1+\varepsilon)A(s,a)) \quad (3)$$

where $r(\theta) = \dfrac{\pi_\theta(a|s)}{\pi_{\theta_k}(a|s)}$, the ratio of the probability under the new and old polices, $A(s, a)$ is the estimated advantage, $\varepsilon$ is a small hyperparameter that limits the new policy close to the old.

Algorithm 1 shows the workflow of the proposed motion planner. It runs the old policy $\pi_{\theta_k}$ in each episode and collects experiences in the training procedure. The policy is updated every $N$ episodes. The target success rate of this motion planner of the new target is $r$. To make sure that navigation targets are in the free space in environments, we set a set of start points including the targets for training the motion planner and new targets just for testing.

| **Algorithm 1** Proposed Motion Planner |
|---|
| 1: Initialize the parameters of DRL model as $\theta_1$ |
|     Set the epochs number $K$, episode number $N$, buffer $B$ |
|     Set a set of target points $P$ in free space, take some out as test set $P^-$ |
|     Set the success rate $r$ |
| 2: **for** $k = 1, 2, …$ **do** |
| 3:    Start the navigation environment, start the robot in the initial position and randomly choose the target position $p_0$ in $P$ |
| 4:    **if** $p_0$ not in $P^-$ **then** |
| 5:      Run policy $\pi_{\theta_k}$ in environment and collect experience $\{s_t, a_t, r_t\}$ to $B$ until terminal |
| 6:      **for** every N episodes in $B$ **then** |
| 7:        Compute estimated advantage $A_t$ |
| 8:        Update $\theta_1$ by formula (3) with $K$ epochs |
| 9:        Remove experiences of these episodes form $B$ |
| 10:     **end for** |
| 11:    **else then** |
| 12:      Run policy $\pi_{\theta_k}$ in environment until terminal and record if arrival |
| 13:      Calculate the success rate $r'$ in the last 100 test episodes |
| 14:      **if** $r' > r$ **then**: |
| 15:        Save the model |
| 16:        **Break** |

| 17. | **end if** |
| 18. | **end if** |
| 19. **end for** | |

*E. Reward Function Definition*

Reward function is crucial for reinforcement learning. Reward function directly guides the learning direction of reinforcement learning. It is an important issue to design a reward function consistent with the goal. In the navigation task, the goal is to arrive at the target position without collision. Thus a positive reward is arranged if mobile robot arrives at the target position and a negative reward is arranged if the robot collides with any obstacles. However, the reward is too sparse to direct the agent to learn just under these two conditions. The robot is not in both conditions in most cases, and it is necessary to provide a reward signal that guides the robot to reach the target position when it moves. The difference in the distance from the target is such a signal. So the reward function is defined as:

$$r(s_t, a_t) = \begin{cases} r_{collision}, & \text{if collision} \\ r_{arrival}, & \text{if } d_t < c_d \\ c_r(d_{t-1} - d_t) - c_p \end{cases} \quad (4)$$

in which $c_p$ is the time penalty for each step to encourage the agent to find the optimal path.

IV. EXPERIMENT

*A. Simulation*

The simulation environments were built for training the motion planner and testing the ability of reasoning. These environments are released as benchmark to compare the navigation algorithms, two of them are shown in Figure 3.

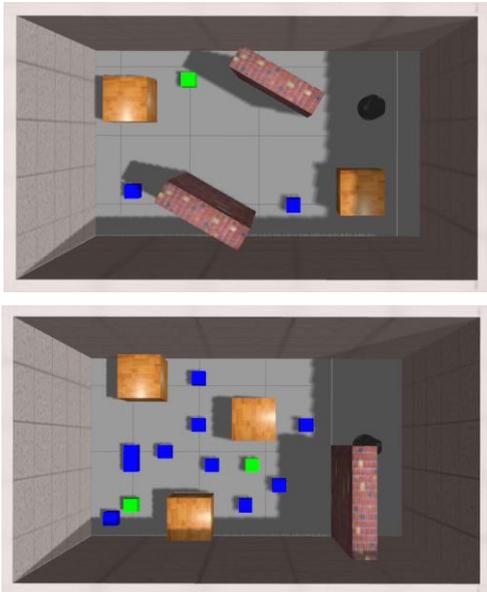

Figure 3.   Navigation environment, *env1*(top) and *env2* (down)

In Figure 3, the position of mobile robot in each indoor environment is its start position. The blue cubes are targets for training the motion planner, experiences to these targets are recorded to update the model. And the green cubes are targets just for evaluating the planner and testing the performance of reasoning. The test target in *env1* is deliberately selected in the area where training targets are seldom, making the robot fully explore. The mobile robot mounted a visual sensor to receive the real-time RGB raw image form the field of view (FOV) of the robot.

*B. Simulation Result and Analysis*

Set the target success rate of the new target is 60%, which is quite a good performance in the second level using RGB images as visual input. The results in *env1* are shown in Figure4.

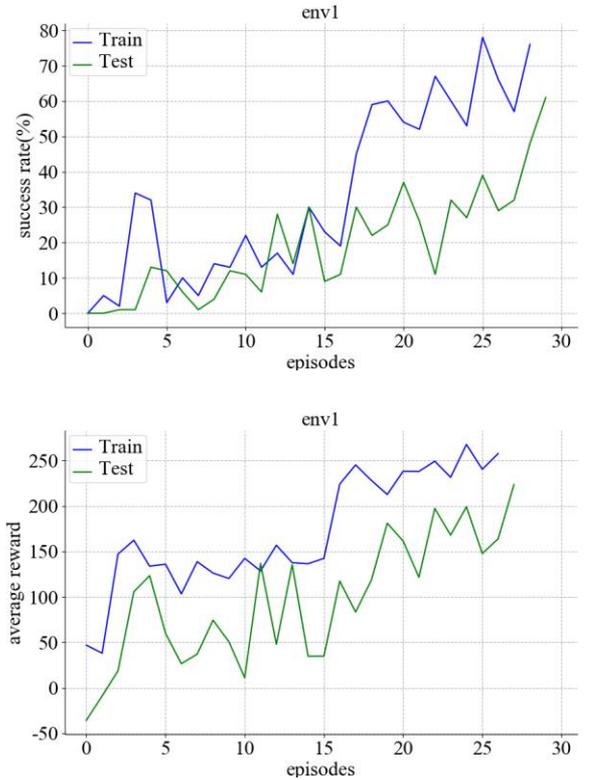

Figure 4.   Success rate and average reward in *env1*

*Success rate* measures the success rate of navigation task of the last 100 episodes. *Average reward* is the average of total rewards of each specified episodes. The specified number for calculating *success rate* and *average reward* is 200 in training and 100 in testing. When the mobile robot can arrive at the new target at a success rate of 60%, it can reach the two positions of training at a success rate of more than 75%.

The results in *env2* are shown in Figure 5. In *env2* the specified number for calculating *success rate* and *average reward* is 500 in training and 100 in testing. *Env2* is more complicated than *env1* and we set many training targets in different free areas to ensure that the robot can fully explore the environment. It's different the method we took in *env1*, in which there is only testing target in some free spaces. This difference makes the performance gap between training and testing in *env2* is less than the performance gap in *env1*.

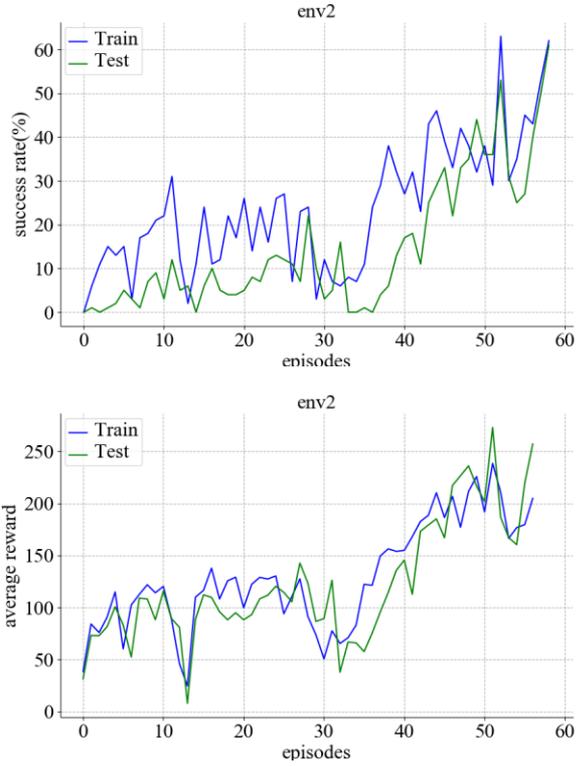

Figure 5.  Success rate and average reward in *env2*

The robot needs to have the ability of reasoning to reach new targets. Specifically, it can infer the depth from visual images and can infer the direction of motion to the target from the relative position. Figure 6 shows a trajectory to a new target in one episode in *env2*. In this trajectory, the robot moved through a circle to adjust its direction so as to reach the target without collision with any obstacles.

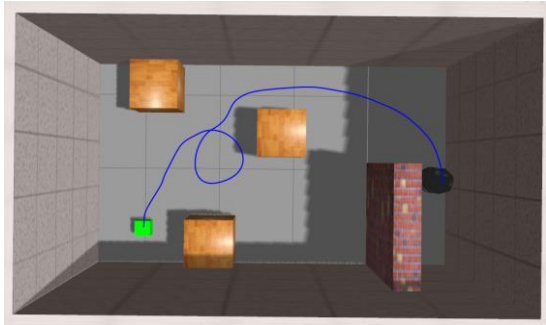

Figure 6.  Reasoning

The proposed motion planner can arrive at new targets in all the built environments with a success rate of more than 60%. Actually the ultimate success rate is 81% in *env1* and 69% in *env2*.

### C. Real Environment

We then implemented the experiment in real environment. The weight trained in *env2* is used as the intial value of the planner. The mobile robot in our lab is shown in Figure 7. It's equipped with ultrasonic sensors which can help the robot stop when it is close to the obstacles. The robot arm and gripper mounted on the mobile base are used for grasping objects.

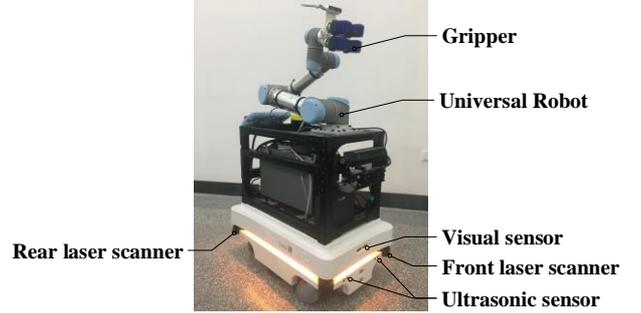

Figure 7.  Mobile robot

We trained the planner in an indoor environment like *env2* with 3000 episodes in three weeks. The planner reached a success rate of 51% in our latest model. The trajectory in one episode to new target is shown in Figure 8. The map built by laser scanner is just for showning.

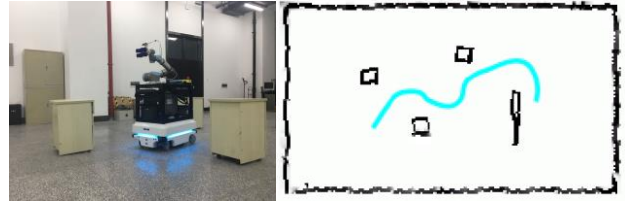

Figure 8.  Environment and Result

## V. CONCLUSION

In this article, we proposed a motion planner that can not only memorize the navigation environment but also reason out the policy to the new target. The stacked LSTM structure is designed to help to remember the environment and learn to build up the internal representations. The motion planner takes RGB image as visual input and odometry as target reference. The velocity and reward in the last timestep can help to understand the navigation task. Experiments were implemented in two different simulation environments and the motion planner got the success rate of more than 60% to new targets. In the experiments, two methods were used to ensure that the robot can fully explore the environment. The built environment is released for future research and the source code is also released to help other researchers to reproduce my results easily. Finally, the experiment in real environment was implemented, and the success rate got 51% by training it in three weeks.

REFERENCES

[1] Stentz A. The focussed D^* algorithm for real-time replanning[C]//IJCAI. 1995, 95: 1652-1659.
[2] Boor V, Overmars M H, Van Der Stappen A F. The Gaussian sampling strategy for probabilistic roadmap planners[C]//ICRA. 1999: 1018-1023.

[3] Anderson P, Chang A, Chaplot D S, et al. On evaluation of embodied navigation agents[J]. arXiv preprint arXiv:1807.06757, 2018.
[4] V. Mnih *et al.*, "Human-level control through deep reinforcement learning," *Nature*, vol. 518, no. 7540, pp. 529–533, Feb. 2015.
[5] D. Silver *et al.*, "Mastering the game of Go without human knowledge," *Nature*, vol. 550, no. 7676, pp. 354–359, Oct. 2017.
[6] T. Haarnoja, V. Pong, A. Zhou, M. Dalal, P. Abbeel, and S. Levine, "Composable Deep Reinforcement Learning for Robotic Manipulation," in *2018 IEEE International Conference on Robotics and Automation (ICRA)*, 2018, pp. 6244–6251.
[7] T. Haarnoja, V. Pong, A. Zhou, M. Dalal, P. Abbeel, and S. Levine, "Composable Deep Reinforcement Learning for Robotic Manipulation," in *2018 IEEE International Conference on Robotics and Automation (ICRA)*, 2018, pp. 6244–6251.
[8] Tai L, Zhang J, Liu M, et al. A survey of deep network solutions for learning control in robotics: From reinforcement to imitation[J]. arXiv preprint arXiv:1612.07139, 2016.
[9] Prioritized experience replay[J]. arXiv preprint arXiv:1511.05952, 2015.
[10] H. van Hasselt, A. Guez, and D. Silver, "Deep Reinforcement Learning with Double Q-Learning," in *Thirtieth AAAI Conference on Artificial Intelligence*, 2016.
[11] Z. Wang, T. Schaul, M. Hessel, H. van Hasselt, M. Lanctot, and N. de Freitas, "Dueling Network Architectures for Deep Reinforcement Learning," *ArXiv151106581 Cs*, Nov. 2015.
[12] M. Hessel *et al.*, "Rainbow: Combining Improvements in Deep Reinforcement Learning," in *Thirty-Second AAAI Conference on Artificial Intelligence*, 2018.
[13] D. Silver, G. Lever, N. Heess, T. Degris, D. Wierstra, and M. Riedmiller, "Deterministic Policy Gradient Algorithms," p. 9.
[14] S. Gu, T. Lillicrap, I. Sutskever, and S. Levine, "Continuous Deep Q-Learning with Model-based Acceleration," p. 10.
[15] J. Schulman, "Trust Region Policy Optimization," p. 9.
[16] Schulman J, Wolski F, Dhariwal P, et al. Proximal policy optimization algorithms[J]. arXiv preprint arXiv:1707.06347, 2017.
[17] Braitenberg V. Vehicles: Experiments in synthetic psychology[M]. MIT press, 1986.
[18] Corke P. Robotics, vision and control: fundamental algorithms In MATLAB® second, completely revised[M]. Springer, 2017.
[19] Dosovitskiy A, Koltun V. Learning to act by predicting the future[J]. arXiv preprint arXiv:1611.01779, 2016.
[20] Zhu Y, Mottaghi R, Kolve E, et al. Target-driven visual navigation in indoor scenes using deep reinforcement learning[C]//2017 IEEE international conference on robotics and automation (ICRA). IEEE, 2017: 3357-3364.
[21] Khan A, Kumar V, Ribeiro A. Learning Sample-Efficient Target Reaching for Mobile Robots[C]//2018 IEEE/RSJ International Conference on Intelligent Robots and Systems (IROS). IEEE, 2018: 3080-3087.
[22] Ma L, Chen J, Liu Y. Using RGB Image as Visual Input for Mapless Robot Navigation[J]. arXiv preprint arXiv:1903.09927, 2019.
[23] Jaderberg M, Mnih V, Czarnecki W M, et al. Reinforcement learning with unsupervised auxiliary tasks[J]. arXiv preprint arXiv:1611.05397, 2016.
[24] Lample G, Chaplot D S. Playing FPS games with deep reinforcement learning[C]//Thirty-First AAAI Conference on Artificial Intelligence. 2017.
[25] P. Mirowski *et al.*, "Learning to Navigate in Complex Environments," *ArXiv161103673 Cs*, Nov. 2016.
[26] Oh J, Chockalingam V, Singh S, et al. Control of memory, active perception, and action in minecraft[J]. arXiv preprint arXiv:1605.09128, 2016.
[27] Gupta S, Davidson J, Levine S, et al. Cognitive mapping and planning for visual navigation[C]//Proceedings of the IEEE Conference on Computer Vision and Pattern Recognition. 2017: 2616-2625.
[28] Gupta S, Fouhey D, Levine S, et al. Unifying map and landmark based representations for visual navigation[J]. arXiv preprint arXiv:1712.08125, 2017.
[29] Savinov N, Dosovitskiy A, Koltun V. Semi-parametric topological memory for navigation[J]. arXiv preprint arXiv:1803.00653, 2018.
[30] Tai L, Paolo G, Liu M. Virtual-to-real deep reinforcement learning: Continuous control of mobile robots for mapless navigation[C]//2017 IEEE/RSJ International Conference on Intelligent Robots and Systems (IROS). IEEE, 2017: 31-36.
[31] Mirowski P, Grimes M, Malinowski M, et al. Learning to navigate in cities without a map[C]//Advances in Neural Information Processing Systems. 2018: 2419-2430.